\documentclass[letterpaper, 10 pt, conference]{ieeeconf}  

\IEEEoverridecommandlockouts                              

\usepackage{graphicx} 
\usepackage[font=small,labelfont=bf]{caption}
\usepackage{float}
\usepackage{amsmath} 
\usepackage{amsfonts,amssymb}
\usepackage[dvipsnames]{xcolor}
\usepackage{algorithm, algorithmic}
\usepackage{tabularx, booktabs, makecell, caption}
\usepackage{tabularray}
\usepackage{siunitx}
\usepackage{pifont}

\usepackage{cite}
\usepackage[most]{tcolorbox}
\usepackage{lipsum} 
\tcbset{
  promptstyle/.style={
    colback=gray!10,    
    colframe=gray!70,   
    fonttitle=\bfseries,
    boxrule=0.5pt,
    arc=2mm,            
    left=2mm,
    right=2mm,
    top=1mm,
    bottom=1mm,
    enhanced,
    breakable           
  }
}

\usepackage[colorlinks=true, linkcolor=blue, citecolor=green, urlcolor=red]{hyperref}

\makeatletter
\def\algbackskip{\hskip-\ALG@thistlm}
\makeatother

\title{\LARGE \bf
INTENTION: Inferring Tendencies of Humanoid Robot Motion Through Interactive Intuition and Grounded VLM
}

\author{Jin Wang$^{1}$$^{*}$, Weijie Wang$^{1}$, Boyuan Deng$^{1}$, Heng Zhang$^{2,3}$, Rui Dai$^{1}$, Nikos Tsagarakis$^{1}$
\thanks{†This work was supported by the European Union’s Horizon 2020 research and innovation programme, euROBIN EPUE034001, and Leonardo Joint Lab JL Leonardo ETCM058501.}
\thanks{$^{1}$Humanoids and Human-Centered Mechatronics (HHCM), Istituto Italiano di Tecnologia, Genoa, Italy.}
\thanks{$^{2}$Human-Robot Interfaces and Interaction Lab, Istituto Italiano di Tecnologia, Genoa, Italy.}
\thanks{$^{3}$Ph.D. program of national interest in Robotics and Intelligent Machines (DRIM) and Università di Genova, Genoa, Italy.}
\thanks{$^{*}$Corresponding author: {\tt\small wang.jin@iit.it}}
}

\begin{document}

\maketitle
\thispagestyle{empty}
\pagestyle{empty}

\begin{abstract}
Traditional control and planning for robotic manipulation heavily rely on precise physical models and predefined action sequences. While effective in structured environments, such approaches often fail in real-world scenarios due to modeling inaccuracies and struggle to generalize to novel tasks. In contrast, humans intuitively interact with their surroundings, demonstrating remarkable adaptability, making efficient decisions through implicit physical understanding. In this work, we propose INTENTION, a novel framework enabling robots with learned interactive intuition and autonomous manipulation in diverse scenarios, by integrating Vision-Language Models (VLMs) based scene reasoning with interaction-driven memory. We introduce Memory Graph to record scenes from previous task interactions which embodies human-like understanding and decision-making about different tasks in real world. Meanwhile, we design an Intuitive Perceptor that extracts physical relations and affordances from visual scenes. Together, these components empower robots to infer appropriate interaction behaviors in new scenes without relying on repetitive instructions.
Videos: \url{https://robo-intention.github.io}
\end{abstract}

\section{INTRODUCTION}

Humans possess an innate ability known as interactive intuition, which allows them to form an intuitive understanding of the physical world through accumulated experience. This arises from continuous interaction with the environment in daily life, enabling humans to quickly judge and reason about interactive properties, such as stability, reachability, and motion trends of objects, without relying on precise physical models. For example, we can easily determine whether a stack of blocks is likely to fall, whether a tool can reach a target, or in which direction an object will move when subjected to an external force. This approximate yet efficient reasoning ability forms the foundation for humans to flexibly handle a wide variety of complex manipulation tasks.

In contrast, traditional robotic control systems often rely on accurate physical modeling and precise environmental perception, typically applied in Task and Motion Planning (TAMP). While these model-based approaches perform well in structured environments such as industrial assembly lines, they often struggle in unstructured settings where perception is uncertain and object properties are unknown. Model-free learning approaches, such as reinforcement learning, offer robustness in some certain scenarios, but are often task-specific. When tasks change, retraining becomes costly, leading to poor flexibility and generalization in new situations.

In recent years, the development of large language models (LLMs) and vision-language models (VLMs) has brought new breakthroughs in robotic cognition and decision-making. These multimodal models exhibit human-level capabilities in semantic understanding, spatial reasoning, and contextual awareness. They can extract scene structure from images and videos, understand object properties and spatial relationships, and infer task intent expressed in human language. This enables robots to autonomously perform navigation and manipulation tasks in unstructured environments, representing a significant step toward embodied intelligence and robotic autonomy.

\begin{figure}
\centerline{\includegraphics[width=7cm]{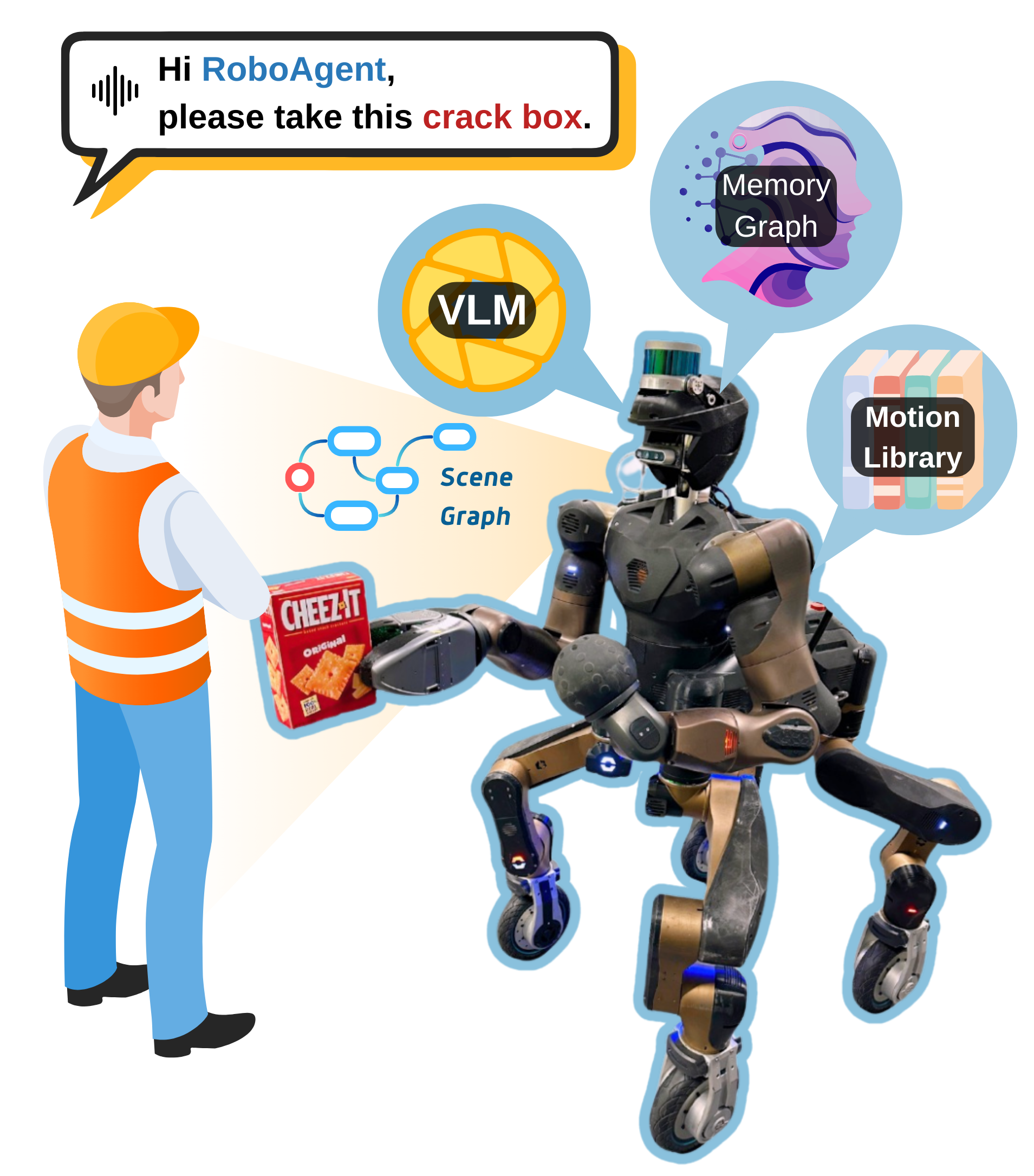}}
\caption{\textbf{INTENTION} enables the humanoid robot to learn, plan, and infer motions to complete manipulation tasks based on intuitive physics and previous experience.}
\label{fig1}
\end{figure}

However, although integrating language models into robotic systems has emerged as a promising direction for enhancing robotic intelligence, there remain significant challenges when applying such approaches to real-world tasks. Existing Vision-Language Action (VLA) methods attempt to directly generate robot actions from pretrained models, enabling robots to interpret task instructions and execute corresponding actions across diverse scenes. Nevertheless, these methods typically rely on large-scale, high-quality datasets for training, which are difficult to obtain-especially for complex, high-degree-of-freedom humanoid robots. Which leads to a limitation on their control accuracy and adaptive capabilities. Moreover, current VLMs used in robotic systems are still primarily focused on static image understanding and lack the ability to model object properties, interactive dynamics, and interactive relationships, which hinders their capacity to handle novel task and predict motion in unstructured environment. In addition, there remains a substantial gap between the high-level semantic outputs of language models and the low-level motion control required for robot execution, particularly in tasks involving whole-body coordination or fine-grained control across multiple articulated components.

To address these challenges, we propose INTENTION, a novel framework that leverages the foundation models in spatial and object-level reasoning to reconstruct a form of physically grounded intuition applicable for real-world humanoid manipulation. To extract task-relevant information from rich interactive environments, we first introduce an \textit{Intuitive Perceptor} based on Vision-Language Models (VLMs). Upon receiving images from the camera, the Perceptor distills spatial and geometric observations and identifies 3D features within the task scene. It then outputs a graph-structured representation encoding both semantic state attributes and geometric relationships among objects and agents. To accumulate knowledge and experience from robot interactions with humans and environment, we further construct a \textit{Memory Graph} (MemoGraph) to support future task and motion planning in novel scenarios. MemoGraph is a topological graph structure that stores semantic information derived from human-robot and environment interactions across diverse task contexts. During each execution, MemoGraph continuously logs the scene information extracted by the Intuitive Perceptor—including semantic instructions, real-world interactive states, and spatial geometry—along with the robot’s action and corresponding feedback. 

These collected representations of real-world physical interactions serve as a rich experiential prior for the robot, enabling it to perform autonomous locomotion and manipulation in unstructured environments. When facing a new task scenario, the agent extracts the current semantic scene state using Perceptor and compares it against previous experiences stored in the MemoGraph, selecting the most relevant interaction behavior that aligns with the present context. With INTENTION, robots are empowered to flexibly adapt to a wide range of tasks across diverse and dynamic environments, while incrementally building a human-level intuitive physics through continual learning from experience. This significantly enhances their autonomy and adaptability in unstructured real-world settings.

Our summary of the main contributions of this work includes:
\begin{itemize}
\item We exploit the integration of grounded language models with intuitive physics and propose INTENTION, a novel framework for autonomous robotic task planning, enabling few-shot training for deployment to humanoids.
\item We propose a VLM-based intuitive perceptor to extract distilled spatial geometry and semantic observation from task scenarios, and further construct a MemoGraph to store the grounded knowledge to guide the behavior selection for diverse tasks combining robotic affordance.
\item We validate the applicability of proposed method through real-world experiments on robotic system, demonstrating effective adaptation across diverse task scenarios. 
\end{itemize}


\section{Related works}

Human interactive intuition refers to the innate ability to infer physical properties and processes from experience without explicit modeling. Endowing robots with such ability refers to studies of cognitive science, machine learning and robotics \cite{groth2021learning}. Priori research \cite{sleiman2021unified, wensing2023optimization, katayama2023model} focus on model-based methods that rely on accurate physical modeling and precise sensing to plan and execute tasks, which perform well in structured environments but struggle with unstructured or dynamic scenarios due to modeling inaccuracies and limited generalization. Recent efforts have explored learning-based methods \cite{ha2024learning} to imitate human-like interactive reasoning. Reinforcement learning~\cite{hengSRL,zhang2025bresa} and imitation learning\cite{parra2023imitation,gams2022manipulation} have been applied to help robots acquire skills from trial-and-error or human demonstrations. Despite progress, many learning-based methods remain task-specific and brittle to environmental changes, requiring extensive retraining for new tasks or scenarios.

Grounding pre-trained language models has emerged as a promising direction in robotics, enabling advanced task planning and reasoning through language. Many recent studies \cite{kawaharazuka2024realworld, li2024robonursevlaroboticscrubnurse, zhang2023lohoravens} leverage large language models (LLMs) for robotic applications, including code generation \cite{liang2023code}, reward shaping \cite{yu2023language, tang2023saytap}, and interactive learning with robots \cite{zha2024distilling, belkhale2024rth, ren2023robots, wang2023prompt}. The integration of multiple modalities—such as vision and audio—has further enhanced the capacity of models to map semantic instructions to perception and behavior \cite{saxena2023multiresolution, lin2023gestureinformed, wang2024autonomous}. Parallel efforts focus on constructing skill libraries \cite{zhang2023bootstrap, ichter2022do, wang2024grounding} that bridge low-level execution with high-level task planning. In addition, recent work \cite{shen2023distilled, huang2023voxposer, lerftogo2023, chen2024spatialvlm} has explored grounding affordances in foundation models to facilitate spatial reasoning and improve manipulation. Despite these advances, the majority of research targets fixed-base robotic arms, with limited exploration into grounding language models for humanoid robots. Some VLA-based approaches \cite{padalkar2023open, geminirobotics2025, kim2024openvla} are capable of generating robot action sequences from multimodal inputs, showing promising results in end-to-end task execution. However, these methods often rely on the collection of large-scale, high-quality datasets to learn robust instruction-action mapping, and tend to struggle with generalization across diverse tasks.

In contrast, our work is the first to explore the use of VLMs for constructing interactive intuition in real-world humanoid manipulation. By leveraging the scene understanding capabilities of VLM, we design an Intuitive Perceptor that extracts semantic and spatial representations from task environments. These are then accumulated into MemoGraph through interaction, enabling the robot to autonomously generate physical behaviors that align with human intuition and understanding when faced with novel tasks and instructions. Moreover, we further investigate through experiments whether robots can infer appropriate actions based solely on the situational context, even under minimal or zero explicit instruction. 

\section{Methodology}

\subsection{Problem Statement} 
In the learning phase, the robot is engaged in a sequence of interaction episodes aimed at constructing a memory-based representation of interactive experiences. We predefined a \textit{skill library} \( \Pi = \{ \pi_1, \pi_2, ..., \pi_n \} \) for the robot, where each skill \( \pi \) denotes a parametrized action or perception primitive executable on the robot hardware. During each interaction \( n \), the robot first observes a scene state \( S_n \), which consists of the human-issued instruction \( I_n \) and the raw sensory input of the environment. An \textit{Intuitive Perceptor} \( P \) processes the scene and extracts the object/environment state \( T_n \), which reflects the spatial and semantic configuration relevant to the task. A vision-language-based planner \( V_p \) then selects the most appropriate action \( a_n \in \Pi \) conditioned on \( I_n \), \( S_n \), and \( T_n \), which is subsequently executed by the robot. After execution, a task success evaluation \( C_n \) is computed to assess the outcome. All of this interaction information—\( S_n \), \( T_n \), \( a_n \), and \( C_n \)—is encoded into a graph representation \( g_n \), which is stored in the \textit{MemoGraph} $\mathcal{M}$. This graph-based memory structure is designed to accumulate reusable experience across different task contexts.

In the inference phase, when the robot is presented with a novel task scenario, it begins by observing the current environment state \( S_t \) and extracting the relevant object configuration \( T_t \) using the Intuitive Perceptor \( P \). It then queries the previously built MemoGraph $\mathcal{M}$, retrieving all stored graph instances \( G = \{g_1, g_2, ..., g_n\} \). A graph matching algorithm compares the current state \( (S_t, T_t) \) with past interaction graphs and selects the most relevant memory graph. Instead of directly reusing the previously stored action, the evaluator will filter the most appropriate actions based on the degree of similarity and the current scenario. This adaptive use of past experience enables the robot to respond flexibly to unseen tasks while grounding its behavior in previously acquired interactive intuition.

The above process is described in Algorithm 1. In this way, the robot is enabled to accumulate interaction experience from human instructions and environment feedback into a structured memory graph, and to utilize this memory to generate context-aware, physically grounded actions in new scenarios.

\begin{algorithm}[t]
\caption{Intuitive Physics Learning and Execution}\label{alg}
\hspace*{\algorithmicindent} \textbf{Given}: VLM planner $V_{p}$, a human instruction $i$, a skill library $\Pi$, Intuitive Perceptor $P$, with the MemoGraph $\mathcal{M}$.
\begin{algorithmic}[2]
\STATE \textbf{// Step 1: Data collection and memory construction}
\STATE{Initialize the state $S$, number of steps $n$}

\FOR{$n = 1$ to $N$}
    \STATE Observe scene state $S_n \gets (I_n, \text{raw sensory input})$
    \STATE $T_n \gets P.\texttt{extract\_object\_state}(S_n)$
    \STATE $a_n \gets V_p(I_n, S_n, T_n, \Pi)$
    \STATE Execute action $a_n$
    \STATE $C_n \gets \texttt{evaluate\_task\_completion}()$
    \STATE $g_n \gets \texttt{encode\_graph}(S_n, T_n, a_n, C_n)$
    \STATE $\mathcal{M}.\texttt{store}(g_n)$
\ENDFOR

\vspace{0.5em}
\STATE \textbf{// Step 2: Inference in New Scenarios}
\STATE Observe new scene $S_t$
\WHILE{$C_t \ne "done"$}
    \STATE $S_t \gets \texttt{perceive\_current\_scene}()$
    \STATE $T_t \gets P.\texttt{extract\_object\_state}(S_t)$
    \STATE $g^* \gets \texttt{encode\_graph}(S_t, T_t)$
    \STATE $G \gets \mathcal{M}.\texttt{retrieve\_all}()$
    \STATE $g^* \gets \texttt{Matching}(G)$
    \STATE $a_t \gets \texttt{Similarity evaluation}$
    \STATE Execute action $a_t$
    \RETURN{Action Executed}
    \STATE$C_t \gets update Status$
\ENDWHILE
\end{algorithmic}
\end{algorithm}

\subsection{INTENTION framework}
\begin{figure*}
 \centerline{\includegraphics[width=16cm]{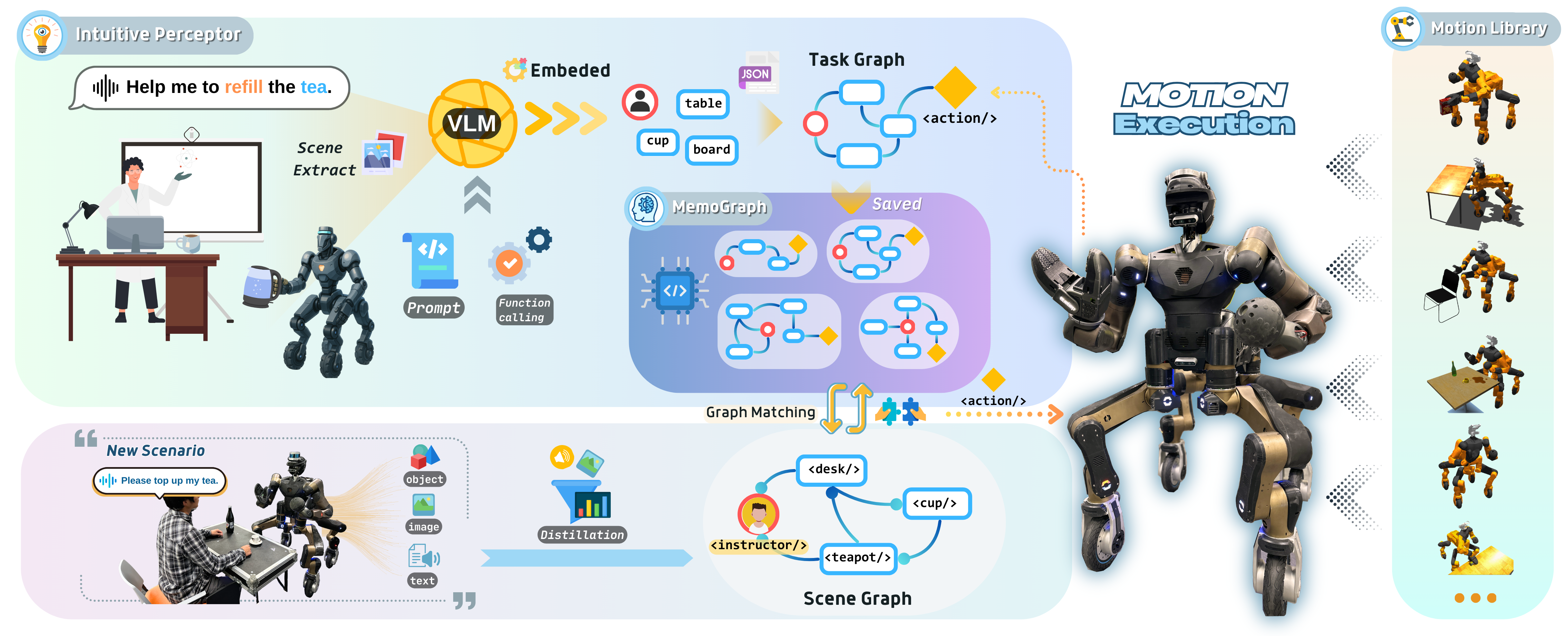}}
\caption{Overview of the Framework. (a) \textbf{Intuitive Perceptor} takes the RGB image and human instruction as input, extracting interactive information of objects and constructs a structured task graph. (b) \textbf{MemoGraph} is responsible for storing knowledge related to the scene of the previous task as well as the actions taken by the robot in the face of different tasks. (c) \textbf{Motion Library} consists of various predefined motion primitives that allow the task planner to choose according to different scenes. When facing with \textbf{New Scenario}, the robot agent will distill the current scene graph and compared it with MemoGraph to select the suitable action.}
\label{framework}   
\end{figure*}

To address the autonomous behavior planning for complex robotics platforms such as humanoid robots and the challenge to reuse human intuition while facing novel tasks and scenarios. This work proposes ‘INTENTION’,  a framework that enables robots with learned interactive intuition and the ability to manipulate in novel scenarios, by integrating VLM-based scene reasoning and interaction-driven memory graph. As shown in Fig.2, we divide the pipeline into four interrelated sectors that are learned and deployed from training to real-world scenarios. The \textit{Intuitive Perceptor} perceives the information of the environment through raw RGB images and grounded visual-language models, and extracts objects state and spatial relationships from the task scenario. The physical scene topology aligns with the robot action for each task execution will be recorded as a task graph and then saved into \textit{MemoGraph}, which serves as a library that stores semantic information derived from human-robot and environment interactions across diverse task contexts. When facing with \textit{new scenario}, the robot first observes the current scene graph and then matches with the task graphs stored in the MemoGraph. The matching results will be evaluated through a language-model based evaluator and to filter the robot motions that best fit the current task scenario. We further construct a robot Motion Library, which includes all the pre-defined motion primitives for different tasks. Each primitive consists of the code interface available for direct control of robot execution and a semantic description of the action.

\subsection{Intuitive Perceptor}

In order to achieve human-level intuitive perception, we utilize the strong scene reasoning capability of a pre-trained visual language model as well as the extensive knowledge of semantic data and propose ‘\textbf{Intuitive Perceptor}’. Although the VLM is able to extract a sufficient amount of corpus information from RGB images, the outputs can be various. In order to obtain the desired answer, we include prompt words in the input and use function calling, where the prompt ensures that the VLM extracts explicit spatial information from the task scene according to the semantic knowledge, and the function definitions enable the output to be generated according to a specific structure and template.

We provide part of the prompts used in Intuitive Perceptor below.

\medskip

\begin{tcolorbox}[promptstyle, fontupper=\ttfamily\small, title=Prompts for Intuitive Perceptor:]
You are an AI agent with human-level commonsense and ability to infer the spatial geometry and relationships in a world scenario. \\ 
You need to extract specific \textcolor{red}{\{object\}} from the provided \textcolor{red}{\{image\}} according to the given \textcolor{red}{\{instruction\}}, and to guess the spatial relationship between \textcolor{red}{\{object1\}} and \textcolor{red}{\{object2\}} in the \textcolor{red}{\{image\}}.  
Answer relationship with one word or phrase. \\
If no \textcolor{red}{\{instruction\}} is detected, then only extract the current scene information.
\end{tcolorbox}

\medskip

The complete prompts and function definitions will be shown on the INTENTION project website.

\subsection{Graph Construction and Matching}
In this subsection, we introduce how to construct a uniform task graph representation for recording scene and robot's action. Then we propose a graph matching algorithm to infer humanoid motion in a new task scenario relying on comparing with previous experience.

\begin{figure*}
 \centerline{\includegraphics[width=18cm]{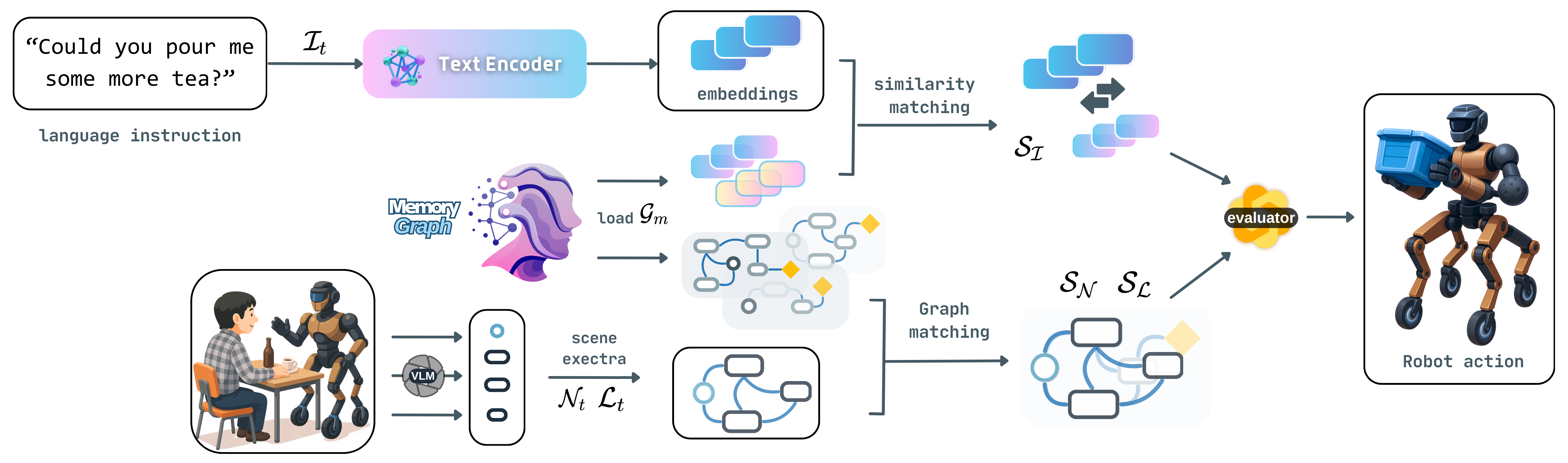}}
\caption{Graph Construction and Matching}
\label{matching}   
\end{figure*}

\subsubsection{MemoGraph}

To leverage the intuitive knowledge and experience acquired from the robot's past interactions with humans and the environment, we design \textbf{MemoGraph}, a graph-based memory structure that stores spatial and geometric observations extracted by the Intuitive Perceptor during task execution. Each task graph in MemoGraph contains semantic instructions, real-world scene states, geometric relations, as well as the robot’s actions and corresponding feedback. This design ensures that the robot can effectively utilize its prior experiences of physical interaction and intuitive understanding to build a systematic framework for cognitive learning and feedback.

We define the task graph $\mathcal{G} = (\mathcal{N}, \mathcal{L}, \mathcal{I})$ as a structured representation consisting of a set of nodes $\mathcal{N}$, a set of links $\mathcal{L}$, and a human instruction $\mathcal{I}$. Each node corresponds to an object or agent (e.g., robot, human), while each link captures the spatial or semantic relationship between two related nodes. The content of nodes, links, and instructions is encoded in text format.

When the robot agent is initialized in a new environment and receives an instruction from the human instructor, the Intuitive Perceptor actively explores the scene and constructs the current scene graph $\mathcal{G}_t$ based on RGB observations.

\subsubsection{Graph Matching}

When the robot agent is presented with a novel task scenario, it queries the constructed MemoGraph $\mathcal{M}$ to retrieve all stored task graph instances, i.e., $\mathcal{M} = {\mathcal{G}_{m_1}, \mathcal{G}_{m_2}, \dots, \mathcal{G}_{m_n}}$. The current scene graph $\mathcal{G}t$ is then matched against each stored task graph $\mathcal{G}_{m}$ in $\mathcal{M}$. To identify the most relevant prior experience and select the robot actions that best align with the current scenario, we propose a scoring-based graph matching method that evaluates the similarity between $\mathcal{G}_t$ and each $\mathcal{G}_{m}$.

Given the current scene graph $\mathcal{G}_t$ and a stored task graph $\mathcal{G}_m$, we design two complementary matching metrics: \textit{instruction similarity} and \textit{graph similarity}, to evaluate how well $\mathcal{G}_t$ matches each graph in the memory.

Formally, for a pair of instructions $\mathcal{I}_t$ and $\mathcal{I}_m$, we first extract their embeddings using an encoder $E(\cdot)$, and compute their similarity using cosine similarity:
\begin{equation}
\mathcal{S_I}(\mathcal{I}_t, \mathcal{I}_m) = \text{cosine-sim}(E(\mathcal{I}_t), E(\mathcal{I}_m))
\end{equation}
where $\mathcal{S}_I$ denotes the instruction similarity score.

To assess the structural similarity between graphs, we compute pairwise similarities between their node sets and link sets. Specifically, we extract embeddings for nodes and links from both graphs, compute the pairwise similarity matrices, and apply thresholding and bipartite matching to determine the best-aligned pairs:
\begin{equation}
\mathcal{P_N} = \mathcal{F}\big(\text{thr}(E(\mathcal{N}_t) \cdot E(\mathcal{N}_m)^\top)\big)
\end{equation}
\begin{equation}
\mathcal{P_N} = \mathcal{F}\big(\text{thr}(E(\mathcal{L}_t) \cdot E(\mathcal{L}_m)^\top)\big)
\end{equation}
Here, $E(\mathcal{N})$ and $E(\mathcal{L})$ denote the embeddings of nodes and links, respectively. $\text{thr}(\cdot)$ is a thresholding function that filters weak similarity scores, and $\mathcal{F}(\cdot)$ represents a bipartite matching function that outputs the optimal assignment between node or link pairs. We also average the similarity matrix of nodes and links to obtain the similarity scores $\mathcal{S_N}$ and $\mathcal{S_L}$.

Then we compute the overall matching score $\mathcal{S_W}$ between the current scene graph and each task graph in the MemoGraph using a weighted combination of the individual similarity scores:
\begin{equation} \mathcal{S_W} = \alpha \cdot \mathcal{S_N} + \beta \cdot \mathcal{S_L} + \gamma \cdot \mathcal{S_I} \end{equation}
where $\alpha + \beta + \gamma = 1$ are weighting coefficients that balance the contributions of node similarity ($\mathcal{S_N}$), link similarity ($\mathcal{S_L}$), and instruction similarity ($\mathcal{S_I}$) during the matching process. All final scores are evaluated by a language-model based evaluator to analyze the plausibility of each match. It retrieved the corresponding action from the task graph of the selected match and then generated the control code for the robot to execute the task.

\section{Experiment and Evaluation}
We proposed real-world experiments to verify the capability of INTENTION by implementing it and assessing its performance on a humanoid robot executing daily manipulation tasks. In this section, we present the details of the experimental setup and design, followed by an analysis and discussion of the results.

\begin{figure}
 \centerline{\includegraphics[width=7cm]{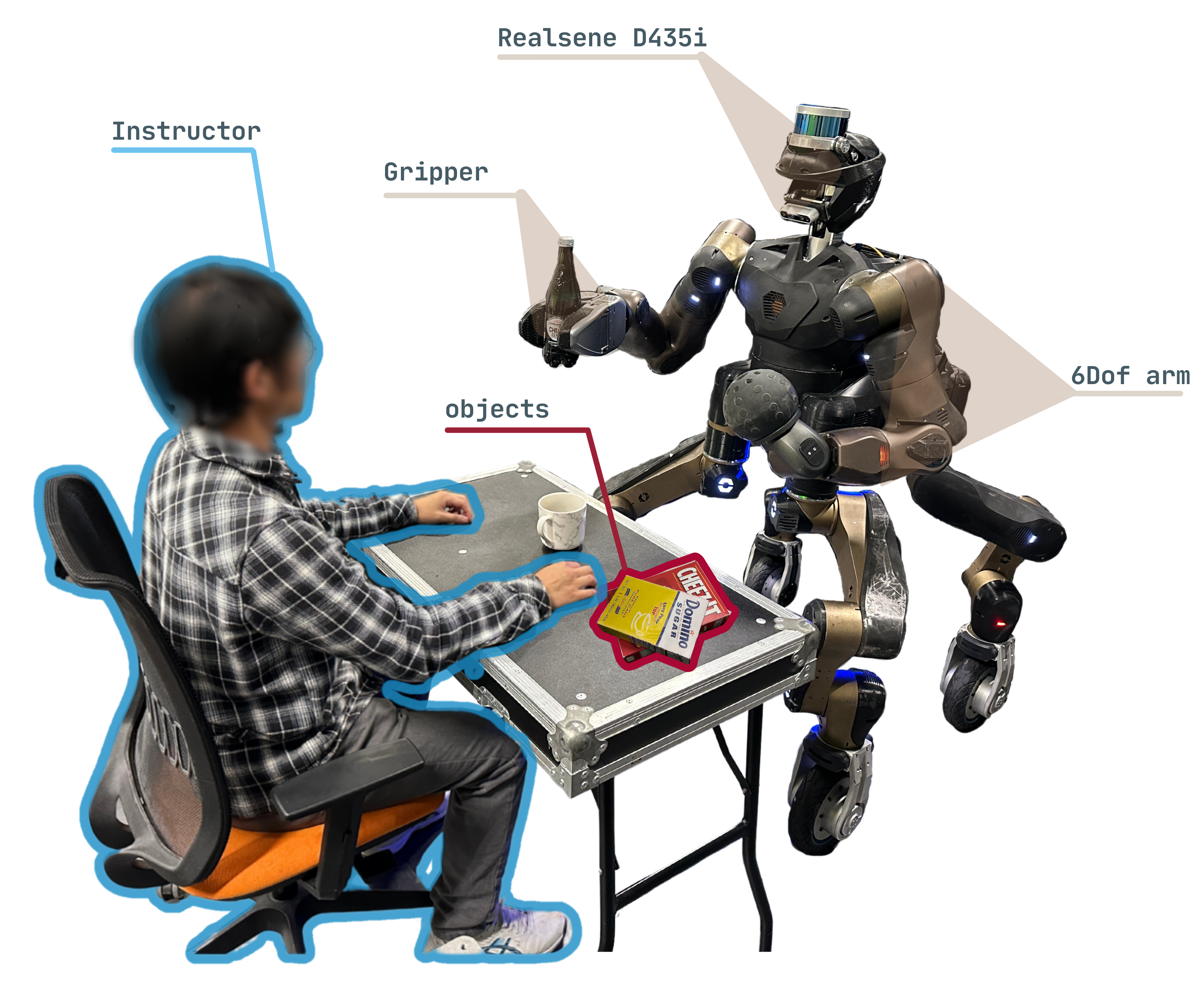}}
\caption{Experiment setup}
\label{setup}   
\end{figure}

\subsection{Experiment Setup}

\begin{table*}[t]
\caption{Comparison of different methods for humanoid manipulation task planning }
\centering
\begin{tblr}{
  cells = {c},
  colspec = {p{1.8cm} p{1cm} p{1cm} p{1.4cm} p{1.4cm} p{0.6cm} p{0.6cm} p{1.5cm} p{0.6cm} p{0.6cm} p{1.6cm}},
  cell{1}{1} = {r=2}{},
  cell{1}{2} = {c=4}{},
  cell{1}{6} = {c=3}{},
  cell{1}{9} = {c=3}{},
  hline{1,7} = {-}{0.08em},
  hline{2} = {2-11}{},
  hline{3-6} = {-}{},
}
\textbf{Method}     & \textbf{Abilities} & & & & \textbf{Manipulation Task} & & & \textbf{interactive Intuition Task} & & \\
           & \small{Autonomy}  & \small{Replan} & \small{Text Input} & \small{Intuition} & $\text{Plan} \uparrow$ & $\text{Succ} \uparrow$ & \scriptsize{Avg. Time $\downarrow$} & $\text{Plan} \uparrow$ & $\text{Succ} \uparrow$ & \scriptsize{Avg. Time $\downarrow$} \\
WB-MPC     & \small{\textit{low}}       & \ding{52}      & \ding{55}          & \ding{55}         & 84\% & 72\% & \scriptsize{35.86s $\pm$ 11.5} & -- & -- & \texttt{TLE} \\
BT-Planner & \small{\textit{low}}       & \ding{55}      & \ding{55}          & \ding{55}         & 92\% & 80\% & \scriptsize{27.40s $\pm$ 4.6} & -- & -- & \texttt{TLE} \\
LLM-BT     & \small{\textit{high}}      & \ding{52}      & \ding{52}          & \ding{55}         & 98\% & 94\% & \scriptsize{22.94s $\pm$ 3.2}  & 20\% & 15\% & \scriptsize{20.65s $\pm$ 8.7} \\
\textbf{INTENTION}  & \small{\textit{high}}      & \ding{52}   & \ding{52}       & \ding{52}      & 96\% & 92\% & \scriptsize{23.75s $\pm$ 4.3} & 84\% & 72\% & \scriptsize{24.86s $\pm$ 2.7}
\end{tblr}
\end{table*}

Our robot is a centaur-like humanoid robot, supported by four legs with wheels. It has 37 degrees of freedom and two arms with one claw gripper on its right arm, enabling it to perform a wide range of manipulation tasks. Equipped with a RealSense Depth camera on its head and torque sensors in the joints throughout its body, the robot possesses extensive perceptual capabilities to measure joint efforts and interaction forces. The robot actions in the motion library were manually designed based on Cartesian I/O without any additional training, and we use Xbot2 to achieve real-time communication between the underlying actuators and the control commands. 

We access the pre-trained \textit{o4-mini} model as the VLM for intuitive perceptor and evaluator from OpenAI API \cite{gpt4v}, and apply function calling to define the structure of the extracted scene graph aligning with the output template using a JSON format. To embed node, link and instruction in the graph, we use CLIP \cite{radford2021learning} as the text encoder. Experiments of the robot performing diverse manipulation task in the real world are demonstrated in the accompanying video.

\subsection{Comparison with other methods}

We conducted both qualitative and quantitative evaluations to compare the proposed INTENTION framework against several existing approaches on our humanoid robot. Specifically, Whole-Body Model Predictive Control (WB-MPC) applies model predictive control across the entire body; BT-Planner relies on manually designed behavior trees (BT) for planning; and LLM-BT \cite{wang2024hypermotion} leverages large language models as high-level planners to generate behavior trees for low-level robotic execution.

The four methods were tested across two types of tasks: (i) a standard manipulation task (Pick and Place) and (ii) a interactive intuition task, where the robot operates without explicit task instructions, relying solely on its embodied physical reasoning capabilities. Experimental results, summarized in Table 1, demonstrate that INTENTION consistently outperforms the baseline methods across multiple capability dimensions. Notably, in interactive intuition tasks that demand a deep understanding of environmental dynamics and affordance extraction, INTENTION exhibits significantly greater adaptability and robustness.

\subsection{Autonomous humanoid motion planning}

\begin{figure}
     \centerline{\includegraphics[width=8cm]{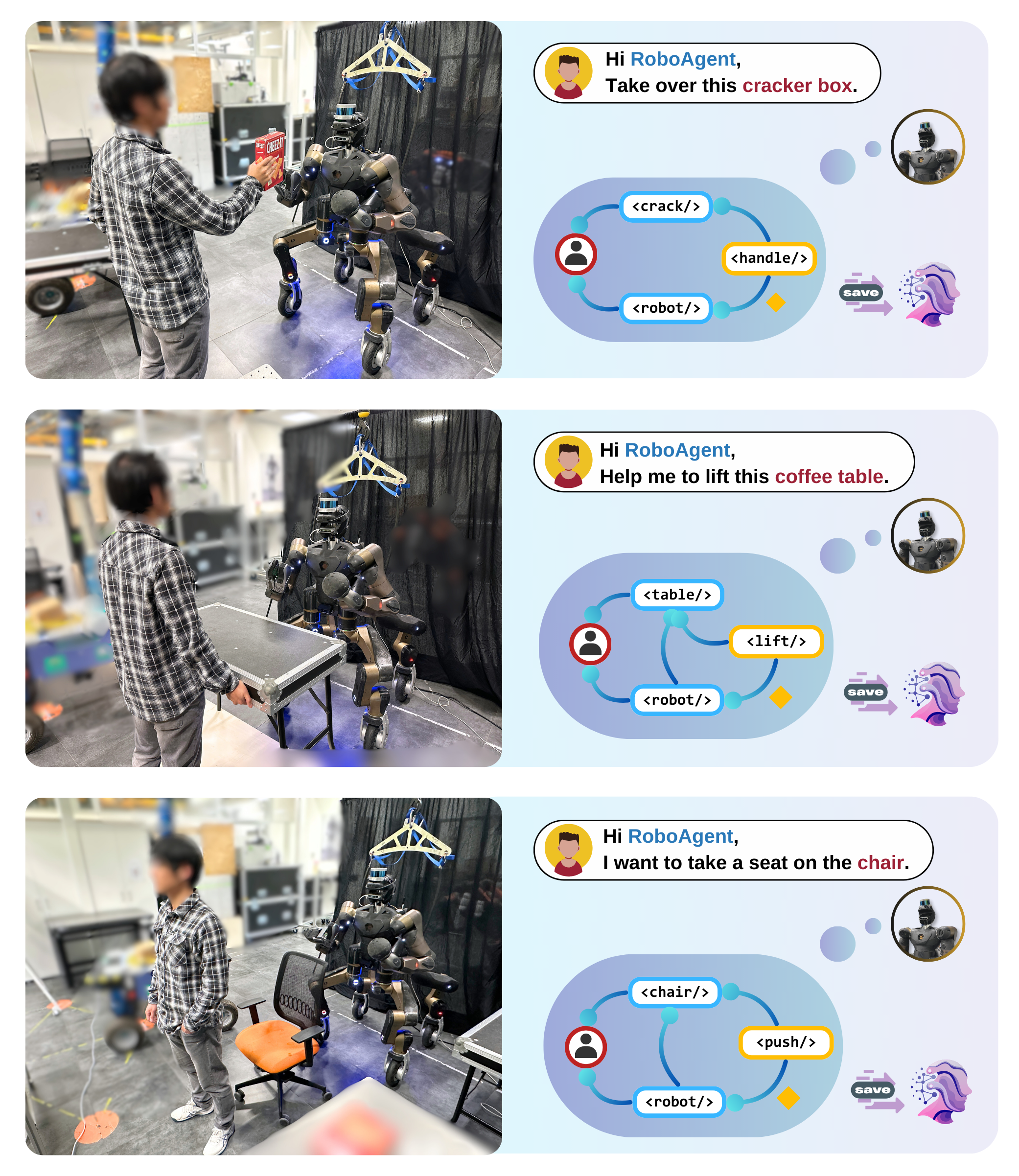}}
    \caption{The process of scenario extraction and MemoGraph construction.}
    \label{simulation}
\end{figure}

\begin{figure}
     \centerline{\includegraphics[width=8cm]{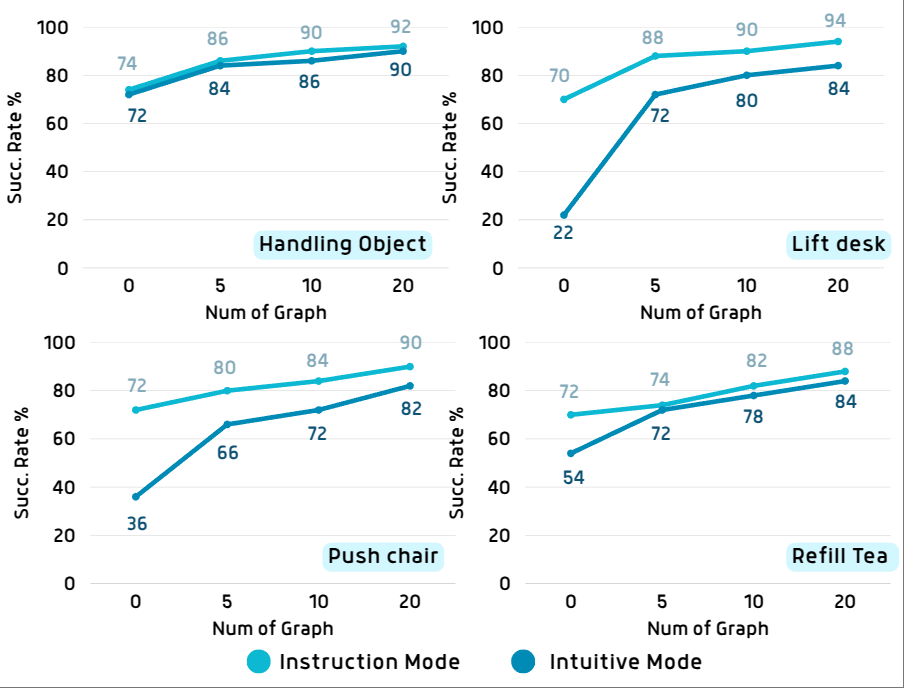}}
    \caption{Task planning success rates under Instruction Mode and Intuitive Mode with different numbers of retrieved graphs.}
    \label{exp_result}
\end{figure}

\begin{figure*}
     \centerline{\includegraphics[width=16cm]{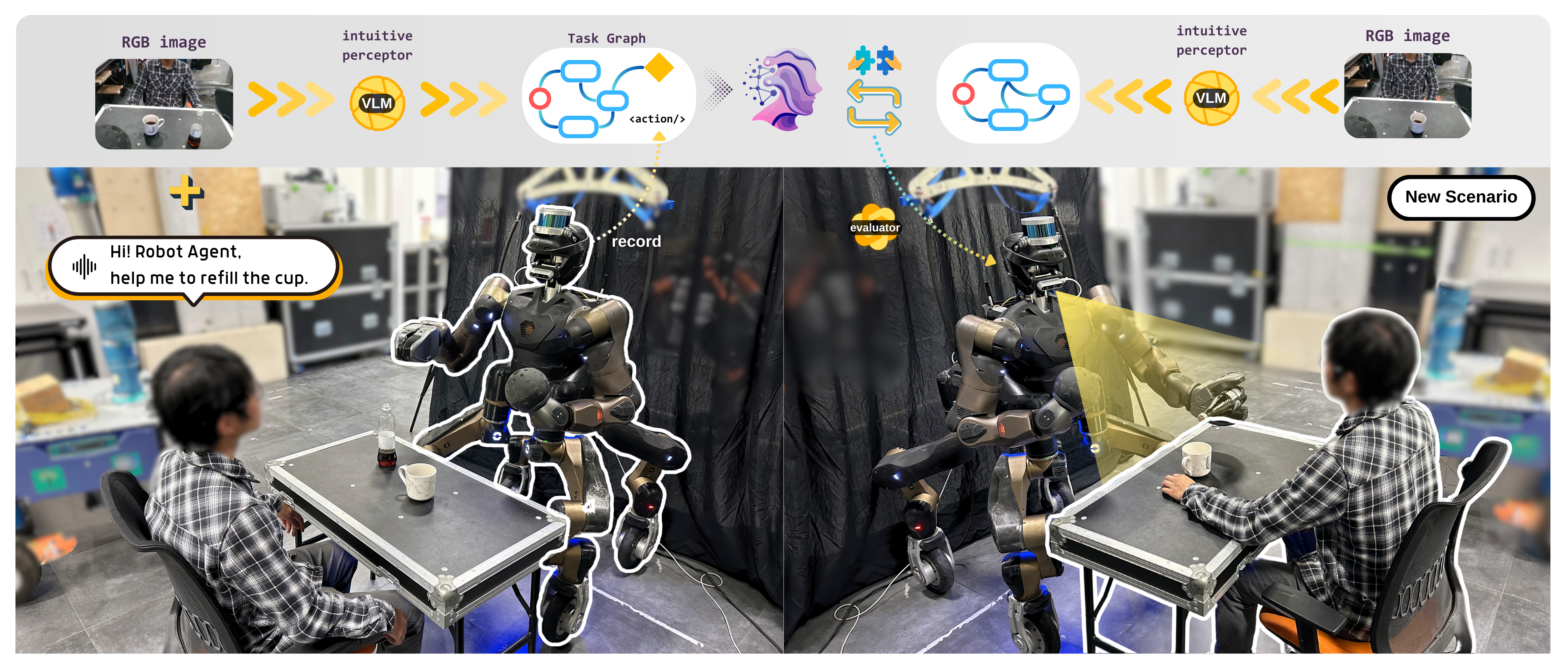}}
    \caption{INTENTION framework pipeline in real-world test, covering scene extraction, task graph generation, MemoGraph construction, and planning/execution for novel task.}
    \label{realworld}
\end{figure*}

\subsubsection{Learning scene extraction and task planning}

To evaluate the scene extraction capability of the VLM-based Intuitive Perceptor (IP) and the contribution of MemoGraph to the robot’s ability to handle novel task scenarios, we conducted a series of experiments on four different manipulation tasks: handling an object, lifting a desk, pushing a chair, and refilling tea.

To construct a sufficiently large MemoGraph dataset, we performed extensive offline experiments across these four tasks. During this process, the IP extracted physical information from each task scene and generated corresponding scene graphs. A task planner based on a VLM was then employed to infer the appropriate actions for the robot based on the given instructions. The execution outcomes were manually evaluated, and the complete task graphs were subsequently stored in MemoGraph, as illustrated in Fig.5.

After building MemoGraph, we further investigated the relationship between the success rate of INTENTION’s scene extraction and task planning, and the number of corresponding task graphs stored in MemoGraph. The experiments involved two different operational modes under novel task scenarios: (i) Instruction Mode, where the robot received explicit semantic instructions to complete the task, and (ii) Intuitive Mode, where no instruction was provided, requiring the robot to rely solely on interactive intuition and prior memory. The experimental results are presented in Fig.6.

\subsubsection{Motion Planning with Intuitive Physics}

After validating the scene extraction and task planning capabilities of INTENTION, we further conducted real-world experiments to assess its ability to infer appropriate robot motions purely based on interactive intuition and prior experience, without relying on any semantic instruction input, across a variety of task scenarios.

In the \texttt{Handling Object} task, the robot is expected to correctly generate a receiving motion when observing the operator handing over different objects. In the \texttt{Lift Desk} task, the robot must infer a collaborative lifting motion based on the operator’s intention to raise the desk. In the \texttt{Push Chair} task, the robot is required to assist the operator in sitting down when it observes the operator approaching and turning their back toward a chair located within the robot’s operational workspace. In the \texttt{Refill Tea} task, the robot must autonomously refill a cup when it observes the operator presenting an empty cup and detects the presence of a teapot either on the table or in the robot’s hand. These tasks pose significant challenges to the robot’s scene understanding and interactive reasoning capabilities.

We conducted 25 trials for each manipulation task and recorded the corresponding success rates, as summarized in Fig.8. Additionally, Fig.7 illustrates the overall workflow of the INTENTION framework, including scene extraction, task graph generation, MemoGraph construction, and the robot agent’s planning and execution process when confronted with novel task scenarios.

\begin{figure}
     \centerline{\includegraphics[width=8cm]{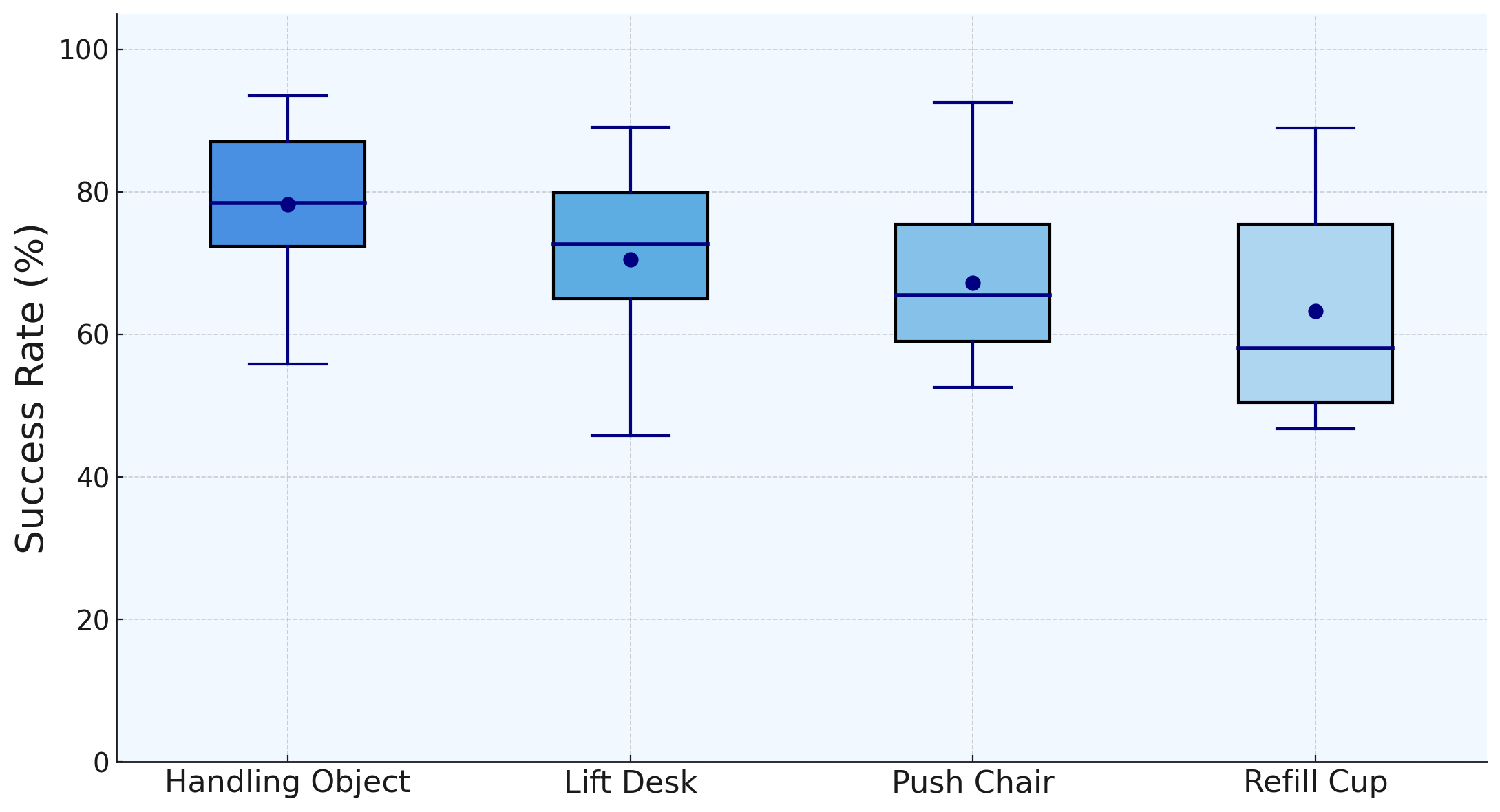}}
    \caption{Success rates of INTENTION across different manipulations relying on interactive intuition}
    \label{exp_result2}
\end{figure}

\subsection{Results analysis}
The experimental results validate the effectiveness of the INTENTION framework across a diverse range of real-world humanoid manipulation tasks.

In the comparison study (Table~I), INTENTION achieves the highest overall performance in both standard manipulation and physical intuition tasks. While methods such as WB-MPC and BT-Planner show moderate success under explicit instructions, they fail to handle tasks requiring physical reasoning without guidance. LLM-BT demonstrates improved autonomy by leveraging language models but struggles in intuitive scenarios, achieving only 15$\%$ success in interactive intuition tasks. In contrast, INTENTION, benefiting from its MemoGraph-based experiential memory and VLM-driven scene understanding, achieves 72$\%$ success in these challenging conditions.

Further analysis of the success rates (Fig.~6 and Fig.~8) reveals that the success probability of INTENTION improves with an increasing number of stored task graphs in MemoGraph, especially for tasks that rely on human-level scene understanding and intuition. Additionally, even without semantic instruction input (Intuitive Mode), INTENTION maintains a high success rate ($\ge 65\%$) across various tasks, demonstrating its ability to infer appropriate robot motions based solely on interactive intuition and prior knowledge. These results collectively indicate that INTENTION significantly enhances the adaptability, robustness, and generalization ability of humanoid robots when facing novel task scenarios.

\section{Conclusion}

In this work, we proposed INTENTION, a novel framework that equips humanoid robots with learned interactive intuition and autonomous task planning capabilities in real-world scenarios. By integrating a VLM-based Intuitive Perceptor for scene understanding and a graph-structured MemoGraph for memory-driven planning, INTENTION enables robots to infer context-appropriate behaviors without relying on explicit instructions. 

Future work will explore scaling MemoGraph to broader task domains, enhancing motion generalization across different robotic platforms, and further improving the efficiency of memory retrieval and action inference under highly dynamic environmental conditions.

\bibliographystyle{IEEEtran}
\bibliography{references}

\addtolength{\textheight}{-12cm}   

\end{document}